\documentclass{article}

\usepackage[final]{neurips_2019}

\usepackage[utf8]{inputenc} 
\usepackage[T1]{fontenc}    
\usepackage{hyperref}       
\usepackage{url}            
\usepackage{booktabs}       
\usepackage{amsfonts}       
\usepackage{nicefrac}       
\usepackage{microtype}      
\usepackage{amsmath}
\usepackage{graphicx}
\usepackage{wrapfig}
\usepackage{mathtools}
\usepackage{tikz}
\usepackage{subfig}
\usepackage[ruled]{algorithm2e}
\usepackage{wrapfig}
\usepackage{lipsum}
\DontPrintSemicolon

\newtheorem{proposition}{Proposition}

\newtheorem{lemma}{Lemma}
\newtheorem{observation}{Observation}

\newcommand{\p}[1]{{#1}^\prime}

\newcommand{\f}[1]{{#1}^{\mathcal{H}}}
\newcommand{\ff}[1]{{#1}^{\mathcal{L}}}
\newcommand{\AC}{\text{DAC}}

\title{$\AC$: The Double Actor-Critic\\Architecture for Learning Options}

\author{
  Shangtong Zhang, Shimon Whiteson \\
  Department of Computer Science\\
  University of Oxford\\
  \texttt{\{shangtong.zhang, shimon.whiteson\}@cs.ox.ac.uk} \\
}

\begin{document}

\maketitle

\begin{abstract}
We reformulate the option framework as two parallel augmented MDPs. 
Under this novel formulation, all policy optimization algorithms can be used off the shelf to learn intra-option policies, option termination conditions, and a master policy over options.
We apply an actor-critic algorithm on each augmented MDP,
yielding the Double Actor-Critic ($\AC$) architecture. 
Furthermore, we show that, when state-value functions are used as critics, one critic can be expressed in terms of the other, and hence only one critic is necessary.
We conduct an empirical study on challenging robot simulation tasks.
In a transfer learning setting, $\AC$ outperforms both its hierarchy-free counterpart and previous gradient-based option learning algorithms.
\end{abstract}

\section{Introduction}
Temporal abstraction (i.e., hierarchy) is a key component in reinforcement learning (RL). 
A good temporal abstraction usually improves exploration \citep{machado2017eigenoption} and enhances the interpretability of agents' behavior \citep{smith2018inference}. 
The option framework \citep{sutton1999between}, which is commonly used to formulate temporal abstraction,
gives rise to two problems: learning options (i.e., temporally extended actions) and learning a master policy (i.e., a policy over options, a.k.a.\ an inter-option policy). 

A Markov Decision Process (MDP, \citealt{puterman2014markov}) with options can be interpreted as a Semi-MDP (SMDP, \citealt{puterman2014markov}), and a master policy is used in this SMDP for option selection.  
While in principle, any SMDP algorithm can be used to learn a master policy, such algorithms are data inefficient
as they cannot update a master policy during option execution.
To address this issue, \cite{sutton1999between} propose \emph{intra-option} algorithms, 
which can update a master policy at every time step during option execution.
Intra-option $Q$-Learning \citep{sutton1999between} is a value-based intra-option algorithm and has enjoyed great success \citep{bacon2017option,riemer2018learning,zhang2018ace}. 

However, in the MDP setting, policy-based methods are often preferred to value-based ones because they can cope better with large action spaces and enjoy better convergence properties with function approximation.
Unfortunately, theoretical study for learning a master policy with policy-based intra-option methods is limited \citep{daniel2016probabilistic,bacon2018thesis} 
and its empirical success has not been witnessed. 
This is the first issue we address in this paper.

Recently, gradient-based option learning algorithms have enjoyed great success \citep{levy2011unified,bacon2017option,smith2018inference,riemer2018learning,zhang2018ace}. 
However, most require algorithms that are customized to the option-based SMDP.  Consequently, we cannot directly leverage recent advances in gradient-based policy optimization from MDPs (e.g., \citealt{schulman2015trust,schulman2017proximal,haarnoja2018soft}). 
This is the second issue we address in this paper.

To address these issues, we reformulate the SMDP of the option framework as two augmented MDPs. 
Under this novel formulation, all policy optimization algorithms can be used for option learning and master policy learning off the shelf and the learning remains intra-option.
We apply an actor-critic algorithm on each augmented MDP,
yielding the Double Actor-Critic (DAC) architecture.
Furthermore, we show that, when state-value functions are used as critics, one critic can be expressed in terms of the other, and hence only one critic is necessary.
Finally, we empirically study the combination of $\AC$ and Proximal Policy Optimization (PPO, \citealt{schulman2017proximal}) in challenging robot simulation tasks. 
In a transfer learning setting, $\AC$+PPO outperforms both its hierarchy-free counterpart, PPO, and previous gradient-based option learning algorithms. 

\section{Background}
\label{sec:bg}
We consider an MDP consisting of a state space $\mathcal{S}$, an action space $\mathcal{A}$, a reward function $r: \mathcal{S} \times \mathcal{A} \rightarrow \mathbb{R}$, a transition kernel $p: \mathcal{S} \times \mathcal{S} \times \mathcal{A}  \rightarrow [0, 1]$, an initial distribution $p_0: \mathcal{S} \rightarrow [0, 1]$ and a discount factor $\gamma \in [0, 1)$. 
We refer to this MDP as $M \doteq (\mathcal{S}, \mathcal{A}, r, p, p_0, \gamma)$ and consider episodic tasks.
In the option framework \citep{sutton1999between},
an option $o$ is a triple of $(\mathcal{I}_o, \pi_o, \beta_o)$, where $\mathcal{I}_o$ is an initiation set indicating where the option can be initiated, $\pi_o : \mathcal{A} \times \mathcal{S} \rightarrow [0, 1]$ is an intra-option policy, and $\beta_o : \mathcal{S} \rightarrow [0, 1]$ is a termination function. 
In this paper, we consider $\mathcal{I}_o \equiv \mathcal{S}$ following \cite{bacon2017option,smith2018inference}. 
We use $\mathcal{O}$ to denote the option set and assume all options are Markov. 
We use $\pi: \mathcal{O} \times \mathcal{S} \rightarrow [0, 1]$ to denote a master policy
and consider the \emph{call-and-return} execution model \citep{sutton1999between}.
Time-indexed capital letters are random variables.
At time step $t$, an agent at state $S_t$ 
either terminates the previous option $O_{t-1}$ w.p. $\beta_{O_{t-1}}(S_t)$ and initiates a new option $O_t$ according to $\pi(\cdot | S_t)$, or proceeds with the previous option $O_{t-1}$ w.p. $1 - \beta_{O_{t-1}}(S_t)$ and sets $O_t \doteq O_{t-1}$.
Then an action $A_t$ is selected according to $\pi_{O_t}(\cdot | S_t)$. The agent gets a reward $R_{t+1}$ satisfying $\mathbb{E}[R_{t+1}] = r(S_t, A_t)$ and proceeds to a new state $S_{t+1}$ according to $p(\cdot | S_t, A_t)$. 
Under this execution model, we have
\begin{align*} 
p(S_{t+1} | S_t, O_t) &= \sum_a \pi_{O_t}(a | S_t) p(S_{t+1} | S_t, a), \\ 
p(O_t | S_t, O_{t-1}) &= (1 - \beta_{O_{t-1}}(S_t))\mathbb{I}_{O_{t-1}=O_t} + \beta_{O_{t-1}}(S_t) \pi(O_t | S_t), \\
p(S_{t+1}, O_{t+1} | S_t, O_t) &= p(S_{t+1}|S_t, O_t)p(O_{t+1}|S_{t+1}, O_t),
\end{align*}
where $\mathbb{I}$ is the indicator function. With a slight abuse of notations, we define $r(s, o) \doteq \sum_a \pi_o(s, a) r(s, a)$. 
The MDP $M$ and the options $\mathcal{O}$ form an SMDP. For each state-option pair $(s, o)$ and an action $a$, we define 
\begin{align*}
\textstyle
q_\pi(s, o, a) \doteq \mathbb{E}_{\pi, \mathcal{O}, p, r}[\sum_{i=1}^\infty \gamma^{i-1} R_{t+i} \mid S_t = s, O_t = o, A_t=a].
\end{align*}
The state-option value of $\pi$ on the SMDP is $q_\pi(s, o) = \sum_a \pi_o(a|s)q_\pi(s, o, a)$.
The state value of $\pi$  on the SMDP is $v_\pi(s) = \sum_o \pi(o | s) q_\pi(s, o)$. They are related as 
\begin{align*}
\textstyle
q_\pi(s, o) = r(s, o) + \gamma \sum_{\p{s}} p(\p{s}| s, o)u_\pi(o, \p{s}), \quad u_\pi(o, \p{s}) = [1 - \beta_o(\p{s})]q_\pi(\p{s}, o) + \beta_o(\p{s})v_\pi(\p{s}),
\end{align*}
where $u_\pi(o, \p{s})$ is the option-value upon arrival \citep{sutton1999between}. 
Correspondingly, we have the optimal master policy $\pi^*$ satisfying $v_{\pi^*}(s) \geq v_\pi(s) \, \forall (s, \pi)$.
We use $q_*$ to denote the state-option value function of $\pi^*$. 


\textbf{Master Policy Learning: } 
To learn the optimal master policy $\pi^*$ given a fixed $\mathcal{O}$, 
one value-based approach is to learn $q_*$ first and derive $\pi^*$ from $q_*$. 
We can use SMDP $Q$-Learning to update an estimate $Q$ for $q_*$ as 
\begin{align*}
\textstyle
Q(S_t, O_t) \leftarrow Q(S_t, O_t) + \alpha \big( \sum_{i=t}^k \gamma^{i-t} R_{i+1} + \gamma^{k-t} \max_o Q(S_{k}, o) - Q(S_t, O_t) \big),
\end{align*}
where we assume the option $O_t$ \emph{initiates} at time $t$ and \emph{terminates} at time $k$ \citep{sutton1999between}. 
Here the option $O_t$ lasts $k - t$ steps. 
However, SMDP $Q$-Learning performs only one single update, yielding significant data inefficiency. 
This is because SMDP algorithms simply interpret the option-based SMDP as a generic SMDP, ignoring the presence of options.
By contrast, \cite{sutton1999between} propose to exploit the fact that the SMDP is generated by options, yielding an update rule:
\begin{align}
\label{eq:intra-q}
\textstyle
Q(S_t, O_t) &\leftarrow Q(S_t, O_t) + \alpha [R_{t+1} + \gamma U(O_t, S_{t+1}) - Q(S_t, O_t) ]  \\
\textstyle
U(O_t, S_{t+1}) &\doteq \big(1 - \beta_{O_t}(S_{t+1}) \big)Q(S_{t+1}, O_t) + \beta_{O_t}(S_{t+1}) \max_o Q(S_{t+1}, o). \nonumber
\end{align}
This update rule is efficient in that it updates $Q$ every time step. 
However, it is still inefficient in that it only updates $Q$ for the executed option $O_t$.
We refer to this property as \emph{on-option}. 
\cite{sutton1999between} further propose Intra-option $Q$-Learning, 
where the update \eqref{eq:intra-q} is applied to \emph{every option} $o$ satisfying $\pi_o(A_t | S_t) > 0$. 
We refer to this property as \emph{off-option}.
Intra-option $Q$-Learning is theoretically justified only when all intra-option policies are deterministic \citep{sutton1999between}. 
The convergence analysis of Intra-option $Q$-Learning with stochastic intra-option policies remains an open problem \citep{sutton1999between}.
The update \eqref{eq:intra-q} and the Intra-option $Q$-Learning can also be applied to off-policy transitions.

\textbf{The Option-Critic Architecture:} 
\cite{bacon2017option} propose a gradient-based option learning algorithm, the Option-Critic (OC) architecture. 
Assuming $\{\pi_o\}_{o \in \mathcal{O}}$ is parameterized by $\nu$ and $\{\beta_o\}_{o \in \mathcal{O}}$ is parameterized by $\phi$,
\cite{bacon2017option} prove that
\begin{align*}
\textstyle
\nabla_\nu v_\pi(S_0) &= \sum_{s, o} \rho(s, o | S_0, O_0) \sum_a q_\pi(s, o, a) \nabla_\nu \pi_o(a | s), \\
\textstyle
\nabla_\phi v_\pi(S_0) &= -\sum_{\p{s}, o} \rho(\p{s}, o | S_1, O_0) \big(q_\pi(\p{s}, o) - v_\pi(\p{s}) \big) \nabla_\phi \beta_o(\p{s}),
\end{align*}
where $\rho$ defined in \cite{bacon2017option} is the unnormalized discounted state-option pair occupancy measure.
OC is \emph{on-option} in that given a transition $(S_t, O_t, A_t, R_{t+1}, S_{t+1})$, it updates only parameters of the executed option $O_t$.
OC provides the gradient for $\{\pi_o, \beta_o\}$ and can be combined with any master policy learning algorithm. 
In particular, \cite{bacon2017option} combine OC with \eqref{eq:intra-q}. 
Hence, in this paper, we use OC to indicate this exact combination.
OC has also been extended to multi-level options \citep{riemer2018learning} and deterministic intra-option policies \citep{zhang2018ace}.

\textbf{Inferred Option Policy Gradient: } 
We assume $\pi$ is parameterized by $\theta$ and define $\xi \doteq \{\theta, \nu, \phi\}$. 
We use $\tau \doteq (S_0, A_0, S_1, A_1, \dots, S_T)$ to denote a trajectory from $\{\pi, \mathcal{O}, M\}$, where $S_T$ is a terminal state. 
We use $r(\tau) \doteq \mathbb{E}[\sum_{t=1}^T \gamma^{t-1} R_t \mid \tau]$ to denote the total expected discounted rewards along $\tau$. 
Our goal is to maximize $J \doteq \int r(\tau)p(\tau)d\tau$. 
\cite{smith2018inference} propose to interpret the options along the trajectory as \emph{latent variables} and marginalize over them when computing $\nabla_\xi J$. 
In the Inferred Option Policy Gradient (IOPG), \cite{smith2018inference} show
\begin{align*}
\textstyle
\nabla_\xi J = \mathbb{E}_\tau[r(\tau) \sum_{t=0}^T \nabla_\xi \log p(A_t | H_t)] = \mathbb{E}_\tau[r(\tau) \sum_{t=0}^T \nabla_\xi \log \big(\sum_o m_t(o) \pi_o(A_t | S_t) \big)],
\end{align*}
where $H_t \doteq (S_0, A_0, \dots, S_{t-1}, A_{t-1}, S_t)$ is the state-action history and $m_t(o) \doteq p(O_t = o|H_t)$ is the probability of occupying an option $o$ at time $t$. 
\cite{smith2018inference} further show that $m_t$ can be expressed recursively via $(m_{t-1}, \{\pi_o, \beta_o \}, \pi)$, allowing efficient computation of $\nabla_\xi J$.
IOPG is an \emph{off-line} algorithm in that it has to wait for a complete trajectory before computing $\nabla_\xi J$. 
To admit \emph{online} updates, \cite{smith2018inference} propose to store $\nabla_\xi m_{t-1}$ at each time step and use the stored $\nabla_\xi m_{t-1}$ for computing $\nabla_\xi m_t$, yielding the Inferred Option Actor Critic (IOAC). 
IOAC is biased in that a stale approximation of $\nabla_\xi m_{t-1}$ is used for computing $\nabla_\xi m_t$. The longer a trajectory is, the more biased the IOAC gradient is. 
IOPG and IOAC are \emph{off-option} in that given a transition $(S_t, O_t, A_t, R_{t+1}, S_{t+1})$, all options contribute to the gradient explicitly. 

\textbf{Augmented Hierarchical Policy:} 
\cite{levy2011unified} propose the Augmented Hierarchical Policy (AHP) architecture.
AHP reformulates the SMDP of the option framework as an augmented MDP. 
The new state space is $\mathcal{S}^\text{AHP} \doteq \mathcal{O} \times \mathcal{S}$.
The new action space is $\mathcal{A}^\text{AHP} \doteq \mathcal{B} \times \mathcal{O} \times \mathcal{A}$, where $\mathcal{B} \doteq \{stop, continue\}$ indicates whether to terminate the previous option or not. 
All policy optimization algorithms can be used to learn an augmented policy 
\begin{align*}
\textstyle
\pi^\text{AHP}\big((B_t, O_t, A_t) | (O_{t-1}, S_t) \big)
\doteq &\pi_{O_t}(A_t | S_t) \Big(\mathbb{I}_{B_t = cont} \mathbb{I}_{O_t = O_{t-1}} + \mathbb{I}_{B_t = stop} \pi(O_t | S_t) \Big) \Big( \\
&\mathbb{I}_{B_t=cont} (1 - \beta_{O_{t-1}}(S_t)) + \mathbb{I}_{B_t = stop} \beta_{O_{t-1}}(S_t) \Big)
\end{align*}
under this new MDP,
which learns $\pi$ and $\{\pi_o, \beta_o\}$ implicitly. 
Here $B_t \in \mathcal{B}$ is a binary random variable.
In the formulation of $\pi^\text{AHP}$, the term $\pi(O_t | S_t)$ is gated by $\mathbb{I}_{B_t = stop}$.
Consequently, the gradient for the master policy $\pi$ is non-zero only when an option terminates (also see Equation~23 in \cite{levy2011unified}). 
This suggests that the master policy learning in AHP is SMDP-style. 
Moreover, as suggested by the term $\pi_{O_t}(A_t | S_t)$ in $\pi^\text{AHP}$, the resulting gradient for an intra-option policy $\pi_o$ is non-zero only when the option $o$ is being executed (also see Equation~24 in \cite{levy2011unified}).
This suggests that the option learning in AHP is \emph{on-option}.
Similar augmented MDP formulation is also used in \cite{daniel2016probabilistic}.

\section{Two Augmented MDPs}
In this section, we reformulate the SMDP as two augmented MDPs: the high-MDP $\f{M}$ and the low-MDP $\ff{M}$. 
The agent makes high-level decisions (i.e., option selection) in $\f{M}$ according to $\pi, \{\beta_o\}$ and thus optimizes $\pi, \{\beta_o\}$.
The agent makes low-level decisions (i.e., action selection) in $\ff{M}$ according to $\{\pi_o\}$ and thus optimizes $\{\pi_o\}$.
Both augmented MDPs share the same samples with the SMDP $\{\mathcal{O}, M\}$.

We first define a dummy option $\#$ and $\mathcal{O}^+ \doteq \mathcal{O} \cup \{\#\}$. 
This dummy option is only for simplifying notations and is never executed.
In the high-MDP, we interpret a state-option pair in the SMDP as a new state and an option in the SMDP as a new action. 
Formally speaking, we define  
\begin{align*}
\f{M} &\doteq \{\f{\mathcal{S}}, \f{\mathcal{A}}, \f{p}, \f{p}_0, \f{r}, \gamma\}, \quad \f{\mathcal{S}} \doteq \mathcal{O}^+ \times \mathcal{S}, \quad \f{\mathcal{A}} \doteq \mathcal{O},\\
\f{p}(\f{S}_{t+1} | \f{S}_t, \f{A}_t) &\doteq \f{p}\big((O_t, S_{t+1}) | (O_{t-1}, S_t), \f{A}_t)\big) \doteq \mathbb{I}_{\f{A}_t = O_t} p(S_{t+1} | S_t, O_t), \\
\f{p}_0(\f{S}_0) \doteq \f{p}_0 \big((O_{-1}, S_0)\big) &\doteq p_0(S_0)\mathbb{I}_{O_{-1} = \#}, \quad
\f{r}(\f{S}_t, \f{A}_t) \doteq \f{r}\big((O_{t-1}, S_t), O_t \big) \doteq r(S_t, O_t)
\end{align*}

We define a Markov policy $\f{\pi}$ on $\f{M}$ as 
\begin{align*}
\f{\pi}(\f{A}_t | \f{S}_t) \doteq \f{\pi}(O_t | (O_{t-1}, S_t)) \doteq p(O_t | S_t, O_{t-1}) \mathbb{I}_{O_{t-1} \neq \#} + \pi(S_t, O_t) \mathbb{I}_{O_{t-1} = \#}
\end{align*}
In the low-MDP, 
we interpret a state-option pair in the SMDP as a new state and leave the action space unchanged. 
Formally speaking, we define
\begin{align*}
\ff{M} &\doteq \{\ff{\mathcal{S}}, \ff{\mathcal{A}}, \ff{p}, \ff{p}_0, \ff{r}, \gamma\}, \quad \ff{\mathcal{S}} \doteq \mathcal{S} \times \mathcal{O}, \quad \ff{\mathcal{A}} \doteq \mathcal{A}, \\
\ff{p}(\ff{S}_{t+1} | \ff{S}_t, \ff{A}_t) &\doteq \ff{p}\big((S_{t+1}, O_{t+1}) | (S_t, O_t), A_t \big) \doteq p(S_{t+1} | S_t, A_t) p(O_{t+1} | S_{t+1}, O_t), \\
\ff{p}_0(\ff{S}_0) \doteq \ff{p} &\big( (S_0, O_0) \big) \doteq p_0(S_0) \pi(S_0, O_0), \quad
\ff{r}(\ff{S}_t, \ff{A}_t) \doteq \ff{r}\big( (S_t, O_t), A_t \big) \doteq r(S_t, A_t)
\end{align*}
We define a Markov policy $\ff{\pi}$ on $\ff{M}$ as
\begin{align*}
\ff{\pi}(\ff{A}_t | \ff{S}_t) \doteq \ff{\pi}\big(A_t | (S_t, O_t) \big) \doteq \pi_{O_t}(A_t | S_t) 
\end{align*}
We consider trajectories with nonzero probabilities and define $\Omega \doteq \{\tau \mid p(\tau | \pi, \mathcal{O}, M) > 0\}$,
$\f{\Omega} \doteq \{\f{\tau} \mid p(\f{\tau} | \f{\pi}, \f{M}) > 0\}$,
$\ff{\Omega} \doteq \{\ff{\tau} \mid p(\ff{\tau} | \ff{\pi}, \ff{M}) > 0\}$. 
With $\tau \doteq (S_0, O_0, S_1, O_1, \dots, S_T)$,
we define a function $\f{f}: \Omega \rightarrow \f{\Omega}$, which maps $\tau$ to $\f{\tau} \doteq (\f{S}_0, \f{A}_0, \f{S}_1, \f{A}_1, \dots, \f{S}_T)$, where $\f{S}_t \doteq (O_{t-1}, S_t), \f{A}_t \doteq O_t, O_{-1}\doteq \#$.
We have:
\begin{lemma}
\label{lem:p1}
$p(\tau | \pi, \mathcal{O}, M) = p(\f{\tau} | \f{\pi}, \f{M})$, $r(\tau) = r(\f{\tau})$, and $\f{f}$ is a bijection.
\end{lemma}
\textit{Proof.} 
See supplementary materials.

We now take action into consideration. 
With $\tau \doteq (S_0, O_0, A_0, S_1, O_1, A_1 \dots, S_T)$,
we define a function $\ff{f}: \Omega \rightarrow \ff{\Omega}$, which maps $\tau$ to $\ff{\tau} \doteq (\ff{S}_0, \ff{A}_0, \ff{S}_1, \ff{A}_1, \dots, \ff{S}_T)$,
where $\ff{S}_t \doteq (S_t, O_t), \ff{A}_t \doteq A_t$.
We have:
\begin{lemma}
\label{lem:p2}
$p(\tau | \pi, \mathcal{O}, M) = p(\ff{\tau} | \ff{\pi}, \ff{M})$, $r(\tau) = r(\ff{\tau})$, and $\ff{f}$ is a bijection.
\end{lemma}
\textit{Proof.} 
See supplementary materials.

\begin{proposition}
\label{pro:p1}
\begin{align*}
\textstyle J \doteq \int r(\tau) p(\tau | \pi, \mathcal{O}, M) d\tau = \int r(\f{\tau}) p(\f{\tau} | \f{\pi}, \f{M}) d \f{\tau} = \int r(\ff{\tau}) p(\ff{\tau} | \ff{\pi}, \ff{M}) d \ff{\tau}.
\end{align*}
\end{proposition}
\textit{Proof.} Follows directly from Lemma~\ref{lem:p1} and Lemma~\ref{lem:p2}.

Lemma~\ref{lem:p1} and Lemma~\ref{lem:p2} indicate that sampling from $\{\pi, \mathcal{O}, M\}$ is equivalent to sampling from $\{\f{\pi}, \f{M}\}$ and $\{\ff{\pi}, \ff{M}\}$. 
Proposition~\ref{pro:p1} indicates that optimizing $\pi, \mathcal{O}$ in $M$ is equivalent to optimizing $\f{\pi}$ in $\f{M}$ and optimizing $\ff{\pi}$ in $\ff{M}$. 
We now make two observations:
\begin{observation}
\label{obs:o1}
$\f{M}$ depends on $\{\pi_o\}$ while $\f{\pi}$ depends on $\pi$ and $\{\beta_o\}$.
\end{observation}
\begin{observation}
\label{obs:o2}
$\ff{M}$ depends on $\pi, \{\beta_o\}$ while $\ff{\pi}$ depends on $\{\pi_o\}$. 
\end{observation}
Observation~\ref{obs:o1} suggests that when we keep the intra-option policies $\{\pi_o\}$ fixed and optimize $\f{\pi}$, we are implicitly optimizing $\pi$ and $\{\beta_o\}$ (i.e., $\theta$ and $\phi$). 
Observation~\ref{obs:o2} suggests that when we keep the master policy $\pi$ and the termination conditions $\{\beta_o\}$ fixed and optimize $\ff{\pi}$, we are implicitly optimizing $\{\pi_o\}$ (i.e., $\nu$). 
All policy optimization algorithms for MDPs can be used off the shelf to optimize the two actors $\f{\pi}$ and $\ff{\pi}$ with samples from $\{\pi, \mathcal{O}, M\}$,
yielding a new family of algorithms for master policy learning and option learning,
which we refer to as the Double Actor-Critic (DAC) architecture.
Theoretically, we should optimize $\f{\pi}$ and $\ff{\pi}$ alternatively with different samples to make sure $\f{M}$ and $\ff{M}$ are stationary.
In practice, optimizing $\f{\pi}$ and $\ff{\pi}$ with the same samples improves data efficiency. 
The pseudocode of DAC is provided in the supplementary materials.
We present a thorough comparison of $\AC$, OC, IOPG and AHP in Table~\ref{tab:cmp}.
$\AC$ combines the advantages of both AHP (i.e., compatibility) and OC (intra-option learning).
Enabling off-option learning of intra-option policies in $\AC$ as IOPG is a possibility for future work.

\begin{table}[h]
\centering
\begin{tabular}{|c|c|c|c|c|}
\hline
     & \multicolumn{1}{c|}{Learning $\pi$} & \multicolumn{1}{c|}{Learning $\{\pi_o, \beta_o\}$} & \multicolumn{1}{c|}{Online Learning} & \multicolumn{1}{c|}{Compatibility} \\ \hline
AHP  & SMDP & on-option  & yes & yes  \\ \hline
OC   & intra-option & on-option & yes & no \\ \hline
IOPG & intra-option & off-option & no & no \\ \hline
$\AC$  & intra-option & on-option & yes & yes \\ \hline
\end{tabular}
\caption{\label{tab:cmp} A comparison of AHP, OC, IOPG and $\AC$. 
(1) For learning $\{\pi_o, \beta_o\}$, all four are intra-option. 
(2) IOAC is online with bias introduced and consumes extra memory.
(3) Compatibility indicates whether a framework can be combined with any policy optimization algorithm off-the-shelf.
}
\end{table}

In general, we need two critics in $\AC$, 
which can be learned via all policy evaluation algorithms.
However, when state value functions are used as critics, 
Proposition~\ref{prop:v} shows that the state value function in the high-MDP ($v_{\f{\pi}}$) can be expressed by the state value function in the low-MDP ($v_{\ff{\pi}}$), and hence only one critic is needed.
\begin{proposition}
\label{prop:v}
$v_{\f{\pi}}\big((o, \p{s})\big) = \sum_{\p{o}} \f{\pi}\big(\p{o} | (o, \p{s}) \big) v_{\ff{\pi}}\big( (\p{s}, \p{o}) \big)$, where
\begin{align*}
\textstyle
v_{\f{\pi}} \big( (o, \p{s}) \big) \doteq \mathbb{E}_{\f{\pi}, \f{M}} [\sum_{i=1}^\infty \gamma^{i-1} \f{R}_{t+i} \mid \f{S}_t = (o, \p{s})], \\
\textstyle
v_{\ff{\pi}}\big( (\p{s}, \p{o}) \big) \doteq \mathbb{E}_{\ff{\pi}, \ff{M}} [\sum_{i=1}^\infty \gamma^{i-1} \ff{R}_{t+i} \mid \ff{S}_t = (\p{s}, \p{o})],
\end{align*}
\end{proposition}
\textit{Proof.} See supplementary materials.

Our two augmented MDP formulation differs from the one augmented MDP formulation in AHP mainly in that we do not need to introduce the binary variable $B_t$.
It is this elimination of $B_t$ that leads to the intra-option master policy learning in DAC and
yields a useful observation: 
the call-and-return execution model (with Markov options) is similar to the polling execution model \citep{dietterich2000hierarchical}, 
where an agent reselects an option \emph{every time step} according to $\f{\pi}$.
This observation opens the possibility for more intra-option master policy learning algorithms.
Note the introduction of $B_t$ is necessary if one would want to formulate the option SMDP as a single augmented MDP and apply standard control methods from the MDP setting. 
Otherwise, the augmented MDP will have $\f{\pi}$ in both the augmented policy and the new transition kernel.
By contrast, in a canonical MDP setting, a policy does not overlap with the transition kernel. 

\textbf{Beyond Intra-option $Q$-Learning: } In terms of learning $\pi$ with a fixed $\mathcal{O}$, Observation~\ref{obs:o1} suggests we optimize $\f{\pi}$ on $\f{M}$. 
This immediately yields a family of policy-based algorithms for learning a master policy,
all of which are intra-option.
Particularly, when we use Off-Policy Expected Policy Gradients (Off-EPG, \citealt{ciosek2017expected}) for optimizing $\f{\pi}$, 
we get all the merits of both Intra-option $Q$-Learning and policy gradients for free.
(1) By definition of $\f{M}$ and $\f{\pi}$, Off-EPG optimizes $\pi$ in an intra-option manner and is as data efficient as Intra-option $Q$-Learning. 
(2) Off-EPG is an off-policy algorithm, so off-policy transitions can also be used, as in Intra-option $Q$-Learning. 
(3) Off-EPG is off-option in that all the options, not only the executed one, explicitly contribute to the policy gradient at every time step.
Particularly, this off-option approach does not require deterministic intra-option policies like Intra-option $Q$-Learning.
(4) Off-EPG uses a policy for decision making, which is more robust than value-based decision making.
We leave an empirical study of this particular application for future work and focus in this paper on the more general problem, learning $\pi$ and $\mathcal{O}$ simultaneously.
When $\mathcal{O}$ is not fixed, the MDP ($\f{M}$) for learning $\pi$ becomes non-stationary. 
We, therefore, prefer on-policy methods to off-policy methods.

\section{Experimental Results}
We design experiments to answer the following questions:
(1) Can $\AC$ outperform existing gradient-based option learning algorithms (e.g., AHP, OC, IOPG)?
(2) Can options learned in $\AC$ translate into a performance boost over its hierarchy-free counterparts?
(3) What options does $\AC$ learn?

$\AC$ can be combined with any policy optimization algorithm, e.g., policy gradient \citep{sutton2000policy}, Natural Actor Critic (NAC, \citealt{peters2008natural}), PPO, Soft Actor Critic \citep{haarnoja2018soft}, Generalized Off-Policy Actor Critic \citep{zhang2019generalized}. 
In this paper, we focus on the combination of $\AC$ and PPO, given the great empirical success of PPO \citep{opeai2018}. 
Our PPO implementation uses the same architecture and hyperparameters reported by \cite{schulman2017proximal}.

\cite{levy2011unified} combine AHP with NAC and present an empirical study on an inverted pendulum domain.
In our experiments, we also combine AHP with PPO for a fair comparison. 
To the best of our knowledge, this is the first time that AHP has been evaluated with state-of-the-art policy optimization algorithms in prevailing deep RL benchmarks. 
We also implemented IOPG and OC as baselines. 
Previously, \cite{klissarov2017learnings} also combines OC and PPO in PPOC.
PPOC updates $\{\pi_o\}$ with a PPO loss and updates $\{\beta_o\}$ in the same manner as OC.
PPOC applies vanilla policy gradients directly to train $\pi$ in an intra-option manner, 
which is not theoretically justified.
We use 4 options for all algorithms, following \cite{smith2018inference}.
We report the online training episode return, smoothed by a sliding window of size 20. 
All curves are averaged over 10 independent runs and shaded regions indicate standard errors. 
All implementations are made publicly available~\footnote{\url{https://github.com/ShangtongZhang/DeepRL}}.
More details about the experiments are provided in the supplementary materials.

\subsection{Single Task Learning}
\label{sec:single}
\begin{figure}[h]
\begin{center}
\includegraphics[width=1\textwidth]{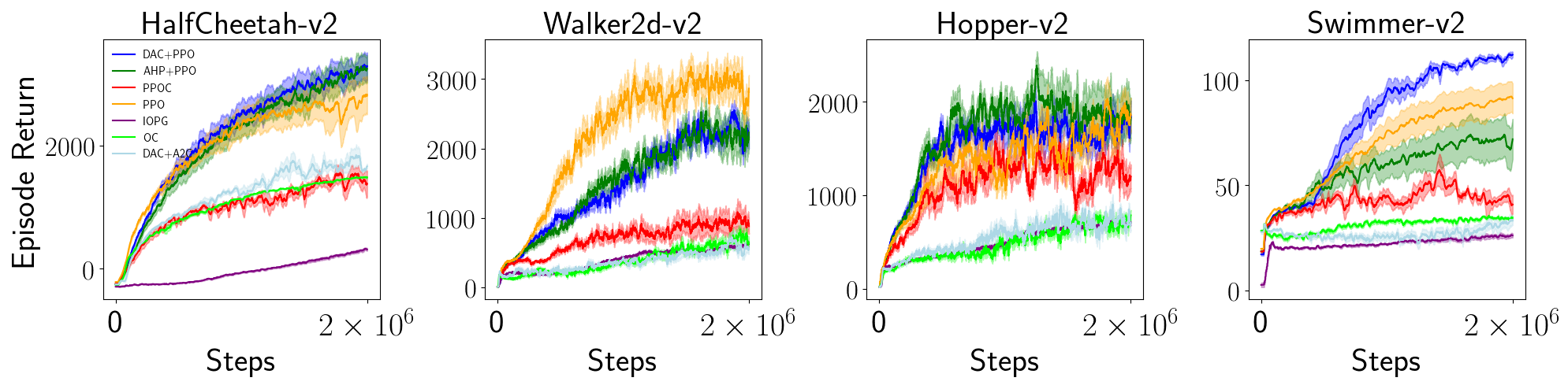}
\end{center}
\caption{\label{fig:single-mean} Online performance on a single task}
\end{figure}

We consider four robot simulation tasks used by \cite{smith2018inference} from OpenAI gym \citep{brockman2016openai}. 
We also include the combination of $\AC$ and A2C \citep{clemente2017efficient} for reference.  
The results are reported in Figure~\ref{fig:single-mean}.

\textbf{Results:} 
(1) Our implementations of OC and IOPG reach similar performance to that reported by \cite{smith2018inference}, 
which is significantly outperformed by both vanilla PPO and option-based PPO (i.e., $\AC$+PPO, AHP+PPO). 
However, the performance of $\AC$+A2C is similar to OC and IOPG. 
These results indicate that the performance boost of $\AC$+PPO and AHP+PPO mainly comes from the more advanced policy optimization algorithm (PPO). 
This is exactly the major advantage of $\AC$ and AHP.
They allow all state-of-the-art policy optimization algorithms to be used off the shelf to  learn options.
(2) The performance of $\AC$+PPO is similar to vanilla PPO in 3 out of 4 tasks. 
$\AC$+PPO outperforms PPO in Swimmer by a large margin. 
This performance similarity between an option-based algorithm and a hierarchy-free algorithm is expected and is also reported by \cite{harb2018waiting,smith2018inference,klissarov2017learnings}.
Within a single task, it is usually hard to translate the automatically discovered options into a performance boost,
as primitive actions are enough to express the optimal policy
and learning the additional structure, the options, may be overhead. 
(3) The performance of $\AC$+PPO is similar to AHP+PPO, as expected.
The main advantage of $\AC$ over AHP is its data efficiency in learning the master policy.
Within a single task, it is possible that an agent focuses on a ``mighty'' option and ignores other specialized options, making master policy learning less important.
By contrast, when we switch tasks, cooperation among different options becomes more important. 
We, therefore, expect that the data efficiency in learning the master policy in $\AC$ translates into a performance boost over AHP in a transfer learning setting.

\subsection{Transfer Learning}
We consider a transfer learning setting where after the first 1M training steps,
we switch to a new task and train the agent for other 1M steps.
The agent is not aware of the task switch.
The two tasks are correlated and we expect learned options from the first task can be used to accelerate learning of the second task. 

\begin{figure}[h]
\begin{center}
\includegraphics[width=0.8\textwidth]{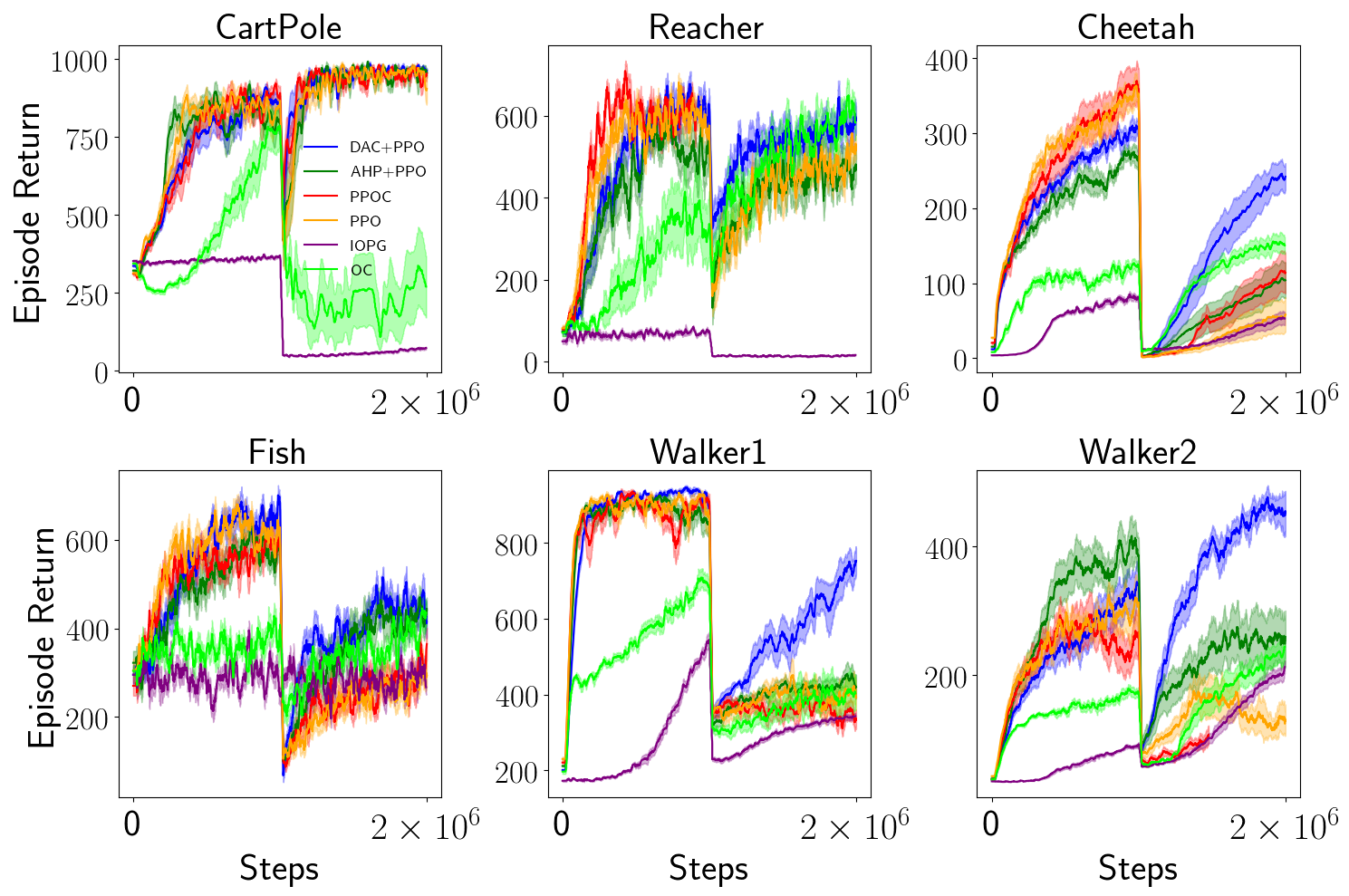}
\end{center}
\caption{\label{fig:transfer-mean} Online performance for transfer learning}
\end{figure}

We use 6 pairs of tasks from DeepMind Control Suite (DMControl, \citealt{tassa2018deepmind}): 
$\texttt{CartPole}=(\texttt{balance}, \texttt{balance\_sparse}), \texttt{Reacher}=(\texttt{easy}, \texttt{hard}), \texttt{Cheetah}=(\texttt{run}, \texttt{backward}), \texttt{Fish}=(\texttt{upright}, \texttt{downleft}), \texttt{Walker1}=(\texttt{squat}, \texttt{stand}), \texttt{Walker2}=(\texttt{walk}, \texttt{backward})$. 
Most of them are provided by DMControl and some of them we constructed similarly as \cite{hafner2018learning}.
The maximum score is always 1000. More details are provided in the supplementary materials.
There are other possible paired tasks in DMControl but we found that in such pairs, PPO hardly learns anything in the second task.
Hence, we omit those pairs from our experiments. The results are reported in Figure~\ref{fig:transfer-mean}.

\textbf{Results:} 
(1) During the first task, $\AC$+PPO consistently outperforms OC and IOPG by a large margin and maintains a similar performance to PPO, PPOC, and AHP+PPO.
These results are consistent with our previous observations in the single task learning setting. 
(2) After the task switch, the advantage of $\AC$+PPO becomes clear.
$\AC$+PPO outperforms all other baselines by a large margin in 3 out of 6 tasks and is among the best algorithms in the other 3 tasks.
This satisfies our previous expectation about $\AC$ and AHP in Section~\ref{sec:single}.
(3) We further study the influence of the number of options in \texttt{Walker2}.
Results are provided in the supplementary materials. 
We find 8 options are slightly better than 4 options and 2 options are worse.
We conjecture that 2 options are not enough for transferring the knowledge from the first task to the second.

\subsection{Option Structures}

\begin{figure}[h]
\begin{center}
\includegraphics[width=0.8\textwidth]{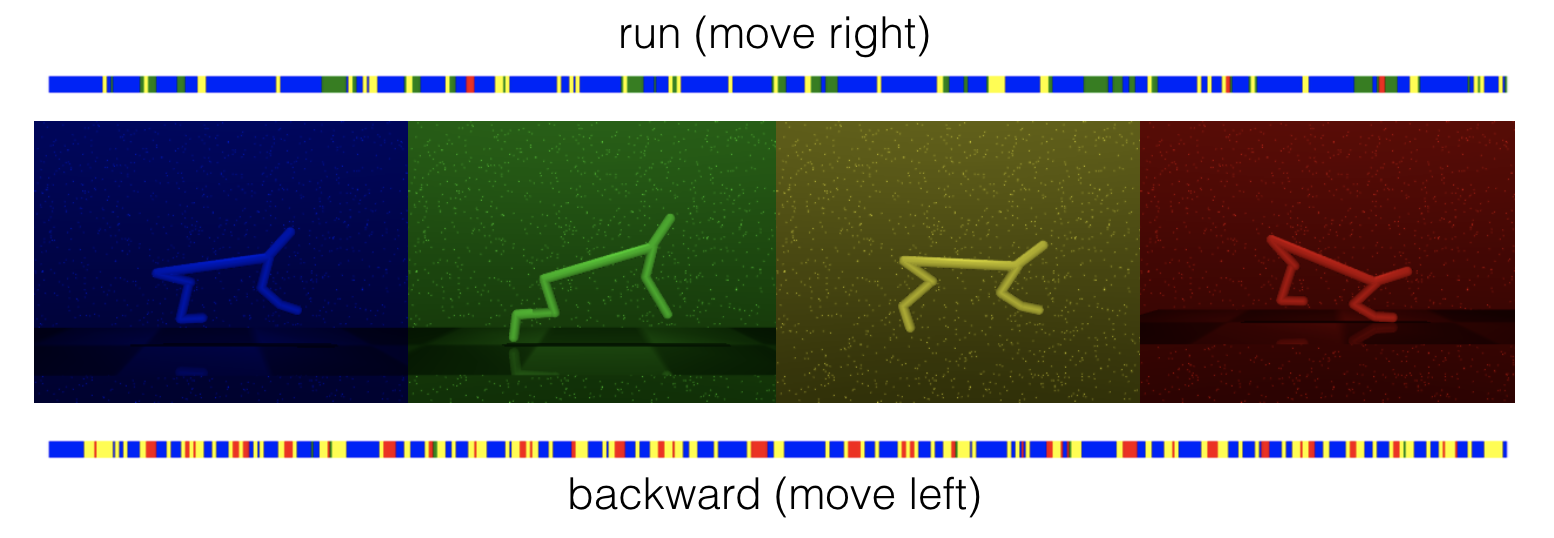}
\end{center}
\caption{\label{fig:vis-options} Learned options and option occupancy of $\AC$+PPO in \texttt{Cheetah}}
\end{figure}

We visualize the learned options and option occupancy of $\AC$+PPO on \texttt{Cheetah} in Figure~\ref{fig:vis-options}. 
There are 4 options in total, displayed via different colors.
The upper strip shows the option occupancy during an episode at the end of the training of the first task (\texttt{run}).
The lower strip shows the option occupancy during an episode at the end of the training of the second task (\texttt{backward}).
Both episodes last 1000 steps.\footnote{The video of the two episodes is available at \url{https://youtu.be/K0ZP-HQtx6M}}
The four options are distinct.
The blue option is mainly used when the cheetah is ``flying''. 
The green option is mainly used when the cheetah pushes its left leg to move right.
The yellow option is mainly used when the cheetah pushes its left leg to move left.
The red option is mainly used when the cheetah pushes its right leg to move left.
During the first task, the red option is rarely used. 
The cheetah uses the green and yellow options for pushing its left leg and uses the blue option for flying. 
The right leg rarely touches the ground during the first episode.
After the task switch, the flying option (blue) transfers to the second task, 
the yellow option specializes for moving left, and the red option is developed for pushing the right leg to the left.

\section{Related Work}

Many components in $\AC$ are not new. 
The idea of an augmented MDP is suggested by \cite{levy2011unified,daniel2016probabilistic}. 
The augmented state spaces $\f{S}$ and $\ff{S}$ are also used by \cite{bacon2017option} to simplify the derivation. 
Applying vanilla policy gradient to $\ff{\pi}$ and $\ff{M}$ leads immediately to the Intra-Option Policy Gradient Theorem \citep{bacon2017option}. 
The augmented policy $\f{\pi}$ is also used by \cite{smith2018inference} to simplify the derivation and 
is discussed in \cite{bacon2018thesis} under the name of mixture distribution.
\cite{bacon2018thesis} discusses two mechanisms for sampling from the mixture distribution: a two-step sampling method (sampling $B_t$ first then $O_t$) and a one-step sampling method (sampling $O_t$ directly), where the latter can be viewed as an expected version of the former. 
The two-step one is implemented by the call-and-return model and is explicitly modelled by \cite{levy2011unified} via introducing $B_t$, which is not used in either \cite{bacon2018thesis} or our work.
\cite{bacon2018thesis} mentions that the one-step modelling can lead to reduced variance compared to the two-step one.
However, there is another significant difference: the one-step modelling is more data efficient than the two-step one.
The two-step one (e.g., AHP) yields SMDP learning while
the one-step one (e.g., our approach) yields intra-option learning (for learning the master policy).
This difference is not recognized in \cite{bacon2018thesis} and we are the first to establish it, both conceptually and experimentally. 
Although the underlying chain of $\AC$ is the same as that of \cite{levy2011unified,daniel2016probabilistic,bacon2017option,bacon2018thesis},
$\AC$ is the first to formulate the two augmented MDPs explicitly.
It is this explicit formulation that allows the off-the-shelf application of all state-of-the-art policy optimizations algorithm and combines advantages from both OC and AHP. 

The gradient of the master policy first appeared in \cite{levy2011unified}.
However, due to the introduction of $B_t$, that gradient is nonzero only if an option terminates.
It is, therefore, SMDP-learning.
The gradient of the master policy later appeared in \cite{daniel2016probabilistic} in the probabilistic inference method for learning options,
which, however, assumes a linear structure and is off-line learning. 
The gradient of the master policy also appeared in \cite{bacon2018thesis}, which is mixed with all other gradients. 
Unless we work on the augmented MDP directly, we cannot easily drop in other policy optimization techniques for learning the master policy, which is our main contribution and is not done by \cite{bacon2018thesis}. 
Furthermore, that policy gradient is never used in \cite{bacon2018thesis}. 
All the empirical study uses Q-Learning for the master policy.
By contrast, our explicit formulation of the two augmented MDPs generates a family of online policy-based intra-option algorithms for master policy learning, 
which are compatible with general function approximation.

Besides gradient-based option learning, there are also other option learning approaches based on finding bottleneck states or subgoals \citep{stolle2002learning,mcgovern2001automatic,silver2012compositional,niekum2011clustering,machado2017laplacian}.
In general, these approaches are expensive in terms of both samples and computation \citep{doina-rlss}.

Besides the option framework, there are also other frameworks to describe hierarchies in RL. 
\cite{dietterich2000hierarchical} decomposes the value function in the original MDP into value functions in smaller MDPs in the MAXQ framework.
\cite{dayan1993feudal} employ multiple managers on different levels for describing a hierarchy.
\cite{vezhnevets2017feudal} further extend this idea to  FeUdal Networks, 
where a manager module sets abstract goals for workers.
This goal-based hierarchy description is also explored by \cite{schmidhuber1993planning,levy2017hierarchical,nachum2018data}.
Moreover, \cite{florensa2017stochastic} use stochastic neural networks for hierarchical RL.
We leave a comparison between the option framework and other hierarchical RL frameworks for future work.

\section{Conclusions}
In this paper, we reformulate the SMDP of the option framework as two augmented MDPs, allowing in an off-the-shelf application of all policy optimization algorithms in option learning and master policy learning in an intra-option manner. 

In $\AC$, there is no clear boundary between option termination functions and the master policy.
They are different internal parts of the augmented policy $\f{\pi}$. 
We observe that the termination probability of the active option becomes high as training progresses, 
although $\f{\pi}$ still selects the same option. 
This is also observed by \cite{bacon2017option}. 
To encourage long options, 
\cite{harb2018waiting} propose a cost model for option switching. 
Including this cost model in $\AC$ is a possibility for future work. 

\subsubsection*{Acknowledgments}
SZ is generously funded by the Engineering and Physical Sciences Research Council (EPSRC). This project has received funding from the European Research Council under the European Union's Horizon 2020 research and innovation programme (grant agreement number 637713). The experiments were made possible by a generous equipment grant from NVIDIA. The authors thank Matthew Smith and Gregory Farquhar for insightful discussions.
The authors also thank anonymous reviewers for their valuable feedbacks.

\medskip

\bibliography{ref.bib}
\bibliographystyle{apa}

\newpage
\appendix

\section{Assumptions and Proofs}
\subsection{Assumptions}
We use standard assumptions \citep{sutton1999between,bacon2017option} about the MDP and options. 
Particularly, we assume all options are Markov.

\subsection{Proof of Lemma~\ref{lem:p1}}
\textit{Proof.}
\begin{align*}
p(\tau | \pi, \mathcal{O}, M) &= p(S_0) \Pi_{t=0}^{T-1}p(O_t | S_t, O_{t-1})p(S_{t+1} | S_t, O_t) \\
 &= p(S_0) \Pi_{t=0}^{T-1}p(O_t | S_t, O_{t-1}) \mathbb{I}_{\f{A}_t = O_t}  p(S_{t+1} | S_t, O_t) \quad \text{(Definition of $\f{A}_t$)} \\
& = \f{p}(\f{S}_0) \Pi_{t=0}^{T-1}\f{\pi}(\f{A}_t | \f{S}_t) \f{p}(\f{S}_{t+1}|\f{S}_t, \f{A}_t) \quad \text{(Definition of $\f{\pi}$ and $\f{p}$)} \\
& = p(\f{\tau} | \f{\pi}, \f{M})
\end{align*}
$r(\tau) = r(\f{\tau})$ follows directly from the definition of $\f{r}$. 
$\f{f}$ is an injection by definition. The definition of $\f{p}$ guarantees $\f{f}$ is a surjection. So $\f{f}$ is a bijection. 

\subsection{Proof of Lemma~\ref{lem:p2}}
\textit{Proof.}
\begin{align*}
p(\tau | \pi, \mathcal{O}, M) &= p(S_0) p(O_0 | S_0) \Pi_{t=0}^{T-1}p(A_t | S_t, O_t) p(S_{t+1} | S_t, A_t) p(O_{t+1}|S_{t+1}, O_t) \\
&= \ff{p}(\ff{S}_0) \Pi_{t=0}^{T-1} \ff{\pi}(\ff{A}_t | \ff{S}_t) \ff{p}(\ff{S}_{t+1} | \ff{S}_t, \ff{A}_t) \quad \text{(Definition of $\ff{\pi}$ and $\ff{p}$)} \\
&= p(\ff{\tau} | \ff{\pi}, \ff{M})
\end{align*}
$r(\tau) = r(\ff{\tau})$ follows directly from the definition of $\ff{r}$.
$\ff{f}$ is an injection by definition.
The definition of $\ff{p}$ guarantees $\ff{f}$ is a surjection.
So $\ff{f}$ is a bijection.

\subsection{Proof of Proposition~\ref{prop:v}}
\textit{Proof.}
\begin{align*}
v_{\f{\pi}} \big( (o, \p{s}) \big) &= \mathbb{E}_{\f{\pi}, \f{M}} [\sum_{i=1}^\infty \gamma^{i-1} \f{R}_{t+i} \mid \f{S}_t = (o, \p{s})] \\
&= \mathbb{E}_{\pi, \mathcal{O}, M} [\sum_{i=1}^\infty \gamma^{i-1} R_{t+i} \mid O_{t-1}=o, S_t = \p{s}] \\
&=u_\pi(o, \p{s}) \quad \text{(Definition of $u_\pi$ in \cite{sutton1999between})} \\
& = [1 - \beta_o(\p{s})]q_\pi(s, o) + \beta_o(\p{s})\sum_{\p{o}}q_\pi(\p{s}, \p{o}) \\
& = \sum_{\p{o}} \f{\pi}\big(\p{o} | (o, \p{s}) \big) q_\pi(\p{s}, \p{o}) \\ 
& = \sum_{\p{o}} \f{\pi}\big(\p{o} | (o, \p{s}) \big) \mathbb{E}_{\pi, \mathcal{O}, M} [\sum_{i=1}^\infty \gamma^{i-1} R_{t+i} \mid S_t = \p{s}, O_t = \p{o}] \\
& = \sum_{\p{o}} \f{\pi}\big(\p{o} | (o, \p{s}) \big) \mathbb{E}_{\ff{\pi}, \ff{M}} [\sum_{i=1}^\infty \gamma^{i-1} \ff{R}_{t+i} \mid \ff{S}_t = (\p{s}, \p{o})] \\
& = \sum_{\p{o}} \f{\pi}\big(\p{o} | (o, \p{s}) \big) v_{\ff{\pi}}\big( (\p{s}, \p{o}) \big)\\
\end{align*}

\section{Details of Experiments}
\subsection{Pseudocode of DAC}
Pseudocode of DAC is provided in Algorithm~\ref{alg:a2c}.
\begin{algorithm}[h]
\textbf{Input:} \;
Parameterized $\pi, \{\pi_o, \beta_o\}_{o \in \mathcal{O}}$ \;
Policy optimization algorithms $\mathbb{A}_1, \mathbb{A}_2$ \; \;

Get an initial state $S_0$ \;
$t \leftarrow 0$ \;
\While{True}{
  Sample $O_t$ from $\f{\pi}\big(\cdot|(O_{t-1}, S_t)\big)$ \;
  Sample $A_t$ from $\ff{\pi}\big(\cdot | (S_t, O_t)\big)$ \;
  Execute $A_t$, get $R_{t+1}, S_{t+1}$ \;
  \tcp{The two optimizations can be done in any order or alternatively}
  Optimize $\f{\pi}$ with $(\f{S}_t, \f{A}_t, R_{t+1}, \f{S}_{t+1})$ and $\mathbb{A}_1$ \;
  Optimize $\ff{\pi}$ with $(\ff{S}_t, \ff{A}_t, R_{t+1}, \ff{S}_{t+1})$ and $\mathbb{A}_2$ \;
  $t \leftarrow t + 1$
}
\caption{\label{alg:a2c} Pseudocode of $\AC$}
\end{algorithm}

\subsection{Details of Environments}
\texttt{CartPole} consists of \texttt{balance} and \texttt{balance\_sparse}, where the latter has a sparse reward. 
\texttt{Reacher} consists of \texttt{easy} and \texttt{hard}, where the latter has a smaller target sphere than the former.
Those four tasks are provided in DMControl.
\texttt{Cheetah} consists of \texttt{run} and \texttt{backward}. 
The former is from DMControl, the latter is from \cite{hafner2018learning}, where the horizontal speed of the cheetah is negated before being used for computing rewards. 
In this task, the cheetah is encouraged to run backward rather than forward.
\texttt{Fish} consists of \texttt{upright} and \texttt{downleft}. 
The former is from DMControl. 
In the latter, we negate the uprightness before using it to compute rewards. 
This task encourages the fish to be ``downleft''.
\texttt{Walker1} consists of \texttt{squat} and \texttt{stand}.
The latter is from DMControl, where a reward is given when the torso height of the walker is larger than $1.2$.
In the former, we give a reward when the torso height is larger than $0.6$.
\texttt{Walker2} consists of \texttt{walk} and \texttt{backward}. 
The former is from DMControl.
In the latter, we negate the horizontal speed as in \texttt{Cheetah-backward}.

\subsection{Implementation Details}
Open AI Gym and DMControl are available at~\url{https://gym.openai.com/} and~\url{https://github.com/deepmind/dm_control}.

\textbf{Function Parameterization:}
We base our parameterization on \cite{schulman2017proximal}.
For an option $o$, $\beta_o$ is parameterized as a two-hidden-layer network. 
A sigmoid activation function is used after the output layer.
$\pi_o$ is parameterized as a two-hidden-layer network.
A Tanh activation function is used after the output layer to output the mean of the Gaussian policy $\pi_o$.
The std of $\pi_o$ is a state-independent variable as \cite{schulman2015trust,schulman2017proximal}.
The master policy $\pi$ is parameterized in the same manner as $\pi_o$ except that a softmax function is used after the output layer.
The value function $q_\pi$ has the same parameterization as $\pi$ 
except that the activation function after the output layer is linear.
All hidden layers have 64 hidden units.
For Mujoco tasks, we use a Tanh activation for hidden layers as suggested by \cite{schulman2017proximal}.
For DMControl tasks, we find a ReLU \citep{nair2010rectified} activation for hidden layers produces better performance.
We use this parameterization for all compared algorithms.

\textbf{Preprocessing:} States are normalized by a running estimation of mean and std.

\textbf{Hyperparameter Tuning:} 
Our PPO implementation is based on the PPO implementation from \cite{baselines}.
Our DAC+PPO and AHP+PPO implementations inherited common hyperparameters from the PPO implementation.
The DAC+A2C implementation is based on the A2C implementation from \cite{baselines}.
The OC and IOPG implementations are based on \cite{smith2018inference} and hyperparameters of A2C from \cite{baselines}.
All the implementations have 4 options following \cite{smith2018inference}.

\textbf{Hyperparameters of PPO:}\\
Optimizer: Adam with $\epsilon=10^{-5}$ and an initial learning rate $3 \times 10^{-4}$ \\
Discount ratio $\gamma$: 0.99 \\
GAE coefficient: 0.95 \\
Gradient clip by norm: 0.5 \\
Rollout length: 2048 environment steps \\
Optimization epochs: 10 \\
Optimization batch size: 64 \\ 
Action probability ratio clip: 0.2 

\textbf{Additional Hyperparameters of DAC+PPO and AHP+PPO:}\\
When optimizing the high MDP, we apply an entropy regularizer with weight $0.01$.
When optimizing the low MDP, we do not use an entropy regularizer.
For DAC+PPO, we perform optimizations for the two MDPs with the same data. For each MDP, we perform 5 optimization epochs. 
So the overall amount of gradient updates remains the same as PPO. 
Our preliminary experiments show that performing the two optimizations alternatively leads to a similar performance.

\textbf{Additional Hyperparameters of PPOC:}\\
Entropy regularizer wieght: 0.01 \\
Option switching penalty weight: 0.01, as suggested by \cite{harb2018waiting}

\textbf{Hyperparameters of DAC+A2C:} \\
Number of workers: 4, as used by \cite{smith2018inference} \\
Optimizer: Adam with $\epsilon=10^{-5}$ and an initial learning rate $3 \times 10^{-4}$ \\
Discount ratio $\gamma$: 0.99 \\
GAE coefficient: 0.95 \\
Gradient clip by norm: 0.5 \\
Entropy regularizer wieght: 0.01 \\
Rollout length: 5 environment steps 

\textbf{Hyperparameters of IOPG:} \\
Number of workers: 4, as used by \cite{smith2018inference} \\
Optimizer: Adam with $\epsilon=10^{-5}$ and an initial learning rate $3 \times 10^{-4}$ \\
Discount ratio $\gamma$: 0.99 \\
Gradient clip by norm: 0.5 

\textbf{Hyperparameters of OC:} \\
Number of workers: 4, as used by \cite{smith2018inference} \\
Optimizer: Adam with $\epsilon=10^{-5}$ and an initial learning rate $3 \times 10^{-4}$ \\
Discount ratio $\gamma$: 0.99 \\
Gradient clip by norm: 0.5 \\
Rollout length: 5 environment steps \\
Entropy regularizer wieght: 0.01 \\
Probability for selecting a random option: 0.1, as used by \cite{harb2018waiting} \\
Target network update frequency: $10^3$ optimization steps \\
Option switching penalty weight: 0.01, as suggested by \cite{harb2018waiting}

\textbf{Computing Infrastructure:} \\
We conducted our experiments on an Nvidia DGX-1 with PyTorch.

\subsection{Other Experimental Results}
\begin{figure}[h]
\begin{center}
\includegraphics[width=0.5\textwidth]{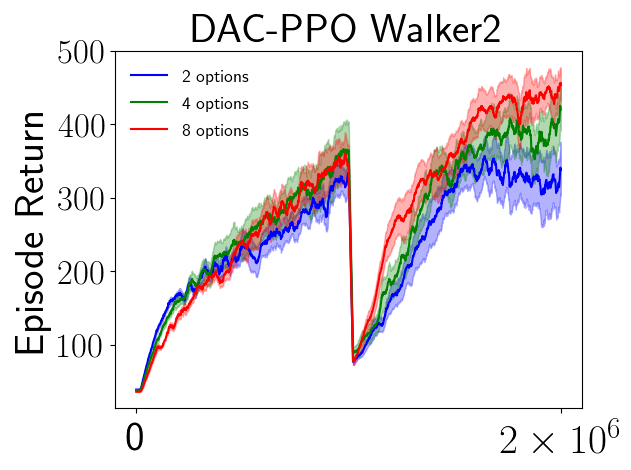}
\end{center}
\caption{\label{fig:ablation} The influence of number of options on performance.}
\end{figure}

Figure~\ref{fig:ablation} studies the influence of the number of options on performance. 
In the first task, the performance is similar. 
In the second task, 8 options are slightly better than 4 options,
while 2 options are clearly worse.

\end{document}